\documentclass[a4paper, 10pt, conference]{ieeeconf}      % Use this line for a4 paper

\IEEEoverridecommandlockouts                              % This command is only needed if 
\usepackage{graphicx}

\title{\LARGE \bf
Hybrid Actuator Design for a Gait Augmentation Wearable
}

\author{Fang Wan$^{1}$, Zheng Wang$^{2}$, Brooke Franchuk$^{3}$, Xinyao Hu$^{4}$, Zhenglong Sun$^{5}$ and Chaoyang Song$^{6}$% <-this % stops a space
\thanks{$^{1}$ Fang Wan is an independent researcher, Wuhan, Hubei, China.}%
\thanks{$^{2}$ Zheng Wang is with the Department of Mechanical and Aerospace Engineering, Monash University, 3800 VIC, Australia.}%
\thanks{$^{3}$ Brooke Franchuk is with the Department of Mechanical and Aerospace Engineering, University of British Columbia, Canada.}%
\thanks{$^{4}$ Xinyao Hu is with the Institute of Human Factors and Ergonomics, Shenzhen University, Shenzhen, Guangdong, China.}%
\thanks{$^{5}$ Zhenglong Sun is with the Institute of Robotics and Intelligent Manufacturing, Chinese University of Hong Kong, Shenzhen, China.}%
\thanks{$^{6,*}$ Chaoyang Song is with the Department of Mechanical and Aerospace Engineering, Monash University, 3800 VIC, Australia. Corresponding Author. 
        {\tt\small songcy@ieee.org}}%
}

\begin{document}

\maketitle
\thispagestyle{empty}
\pagestyle{empty}

%%%%%%%%%%%%%%%%%%%%%%%%%%%%%%%%%%
\begin{abstract}
%%%%%%%%%%%%%%%%%%%%%%%%%%%%%%%%%%

We describe a fluidic actuator design that replaces the sealed chamber of a hydraulic cylinder using a soft actuator to provide compliant linear compression with a large force ($\geq$100 N) at a low operation pressure ($\leq$50 kPa) for lower-limb wearable. The external shells constrain the deformation of the soft actuator under fluidic pressurization. This enables us to use latex party balloons as a quick and cheap alternative for initial design investigation. We found that the forces exerted by the soft material deformation are well-captured by the rigid shells, removing the necessity of explicitly describing the mechanics of the soft material deformation and its interaction with the rigid structure. One can use the classical Force, Pressure and Area formula factored with an efficiency parameter to characterize the actuator performance. Furthermore, we proposed an engineering design of the hybrid actuator using a customized soft actuator placed inside a single shell cavity with an open end for compression force. Our results show that the proposed design can generate a very high force within a short stroke distance. At a low input pressure of 50 kPa, the exerted block force is approaching only about 3\% less than the classical equation predicted. The actuator is fitted to a new gait augmentation design for correcting knee alignment, which is usually challenging for actuators made from the purely soft material. 

\end{abstract}

%%%%%%%%%%%%%%%%%%%%%%%%%%%%%%%%%%
\section{Introduction}
%%%%%%%%%%%%%%%%%%%%%%%%%%%%%%%%%%

Soft robotics is an emerging area in robot design by utilizing material elasticity for compliant, light-weight, and customizable actuation during human-robot interactions, making it a superior choice of design for wearable devices \cite{Rus15}. To stimulate soft material deformation, fluidic pressurization is usually adopted for actuation, leading to the majority of soft robot designs that involves a chamber wrapped by specifically designed soft material for programmable motions. This poses a major challenge in engineering design, which were mainly built on the assumption of rigid mechanics. Soft robot design need to account the soft matter for both motion actuation and power transmission with a changing form factor. This present a major drawback in many current soft robots which can only product a relatively small force or torque, which can only be used for upper limb or hand wearables \cite{Polygerinos13, Yap15, Polygerinos15, Yi16, Chen17}.

The majority of current soft actuators are presented with a relatively long shape with a regular cross-section, generating force under 10 N between a relatively high operating pressure between 200 kPa and 600 kPa (\cite{Polygerinos13, Yap15, Polygerinos15}). Such constrained output, on the other hand, limited the application of such actuators in mainly the hand wearable as assistive gloves. There are a few soft actuator designs aim at overcoming such limitations. For example, the soft actuator designed in \cite{Agarwal16} presented with a high force output of 14 N at 40 kPa input pressure, where a layer of thin, patterned, and unstretchable material is used to wrap around the soft cylinder chamber for mechanically programmable motion generation. In a recent work in \cite{Zhou17}, another two-stage soft actuator design is presented which is capable of grasping 40 N payload in a three-finger configuration. 

\begin{figure}[thpb]
\centering
\includegraphics[width=0.5\textwidth]{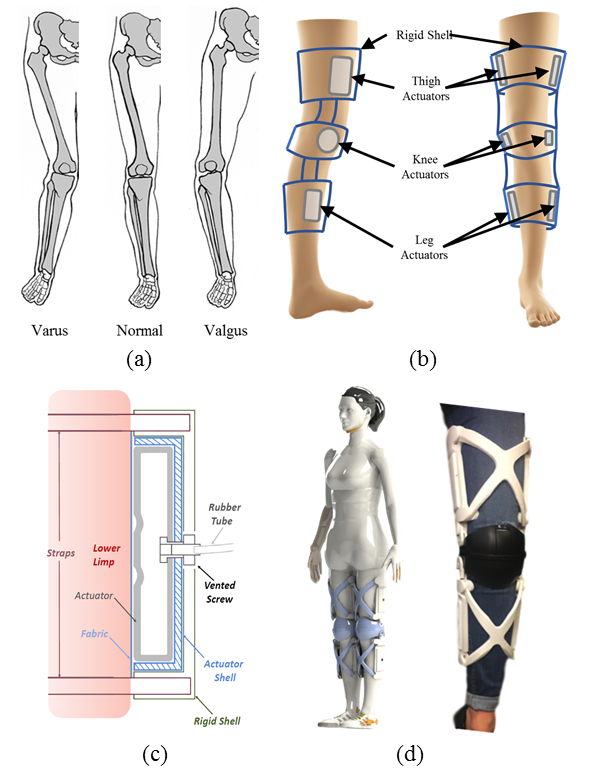}
\caption{Conceptual design of a wearable on the knees for gait augmentation using multiple hybrid actuators made from soft bladders enclosed inside 3D printed rigid cylinder shells for selective compression: (a) misalignment of the knee; (b) conceptual design of the gait augmentation wearable; (c) proposed hybrid actuator design compress under fluidic actuation; and (d) 3D printed wearable prototype.}
\label{fig:PaperOverview}
\end{figure}

The fluidic soft actuators share many similarities with traditional fluidic actuators, such as the hydraulic cylinders. However, a significant amount of pressure is dissipated by the deformed material, especially those made from purely soft material. It has been established that reinforcement by inelastic material can significantly improve the engineering performance of the soft actuator \cite{Polygerinos15,Zhou17,Agarwal16}. There is a research gap in an optimized design methodology between the utilization of fluidic pressurization and the use of soft, elastic material for a light-weight, safe, yet powerful fluidic actuator, which can be bridged by the involvement of material with different elasticity properties. 

In this paper, we propose a fluidic actuator design that replaces the sealed chamber of a hydraulic cylinder using a soft actuator to generate compliant, powerful and light-weight actuation for lower-limb assistance, as shown in Fig. \ref{fig:PaperOverview}, the soft actuator inside the rigid shells seals the pressurized fluids even under a high pressure. Once pressurized, the soft actuator can interact with human body through direct interaction. The rigid shell constrains and guide the soft actuator to deform in a desirable way, which significantly improves to use of fluidic pressure. Effectively, one can describe the overall performance of the actuator by examining the output from the rigid components without specific modeling of the soft matter. Our experiment results shows that the resultant actuator remains efficient in delivering the hydraulic pressuring into actuation force in a compact form factor and a relatively lower input pressure. We further implement our proposed hybrid actuator design into a wearable device for the lower limb knee augmentation through selective actuation of multiple actuators on two sides of the leg in a light-weight design.

%%%%%%%%%%%%%%%%%%%%%%%%%%%%%%%%%%
\section{Experiment Design and Analysis}
%%%%%%%%%%%%%%%%%%%%%%%%%%%%%%%%%%

\begin{figure}[thpb]
\centering
\includegraphics[width=0.5\textwidth]{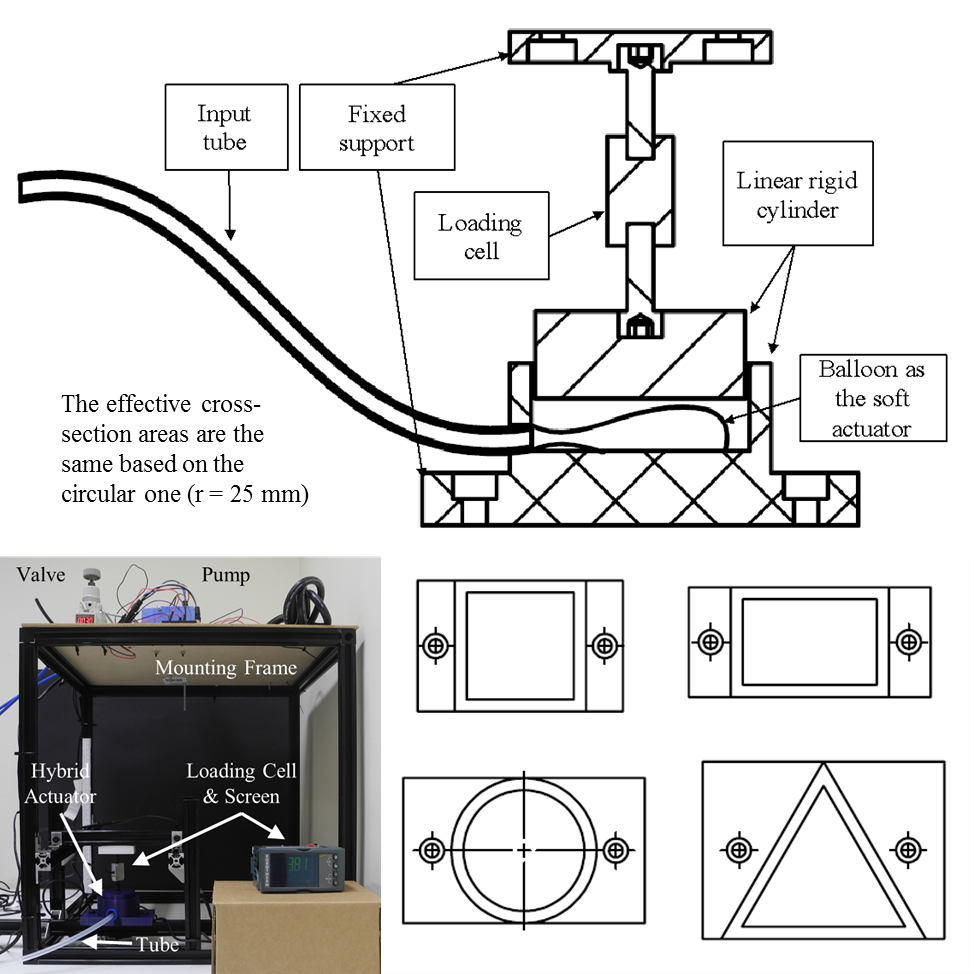}
\caption{Conceptual hybrid actuator design validation using a balloon placed inside 3D printed shells with different geometries but the same areas, including circular, triangle, square and rectangle shapes.}
\label{fig:BalloonExp}
\end{figure}

We used a common latex party balloon instead of a customized, molded actuator to investigate the working mechanism of the proposed actuator design. As shown in Fig. \ref{fig:BalloonExp}, a small latex party balloon is placed inside the chamber of the sliding cylinders. The lower one is mounted on the test rig and the upper moving one is attached to a loading cell, which is also fixed on the test rig. Different cross-section shapes. The balloon is loosely placed inside. After pressurization, the balloon is inflated to fill the space inside the cylinder with a tendency to push the sliding cylinder on the top to move outwards. Under a limited stroke of 5 mm, the balloon is quickly inflated to fill all the cavity inside the rigid shells at a low pressure and starts to push the sliding part against the loading cell, generating force. 

\begin{figure}[thpb]
\centering
\includegraphics[width=0.4\textwidth]{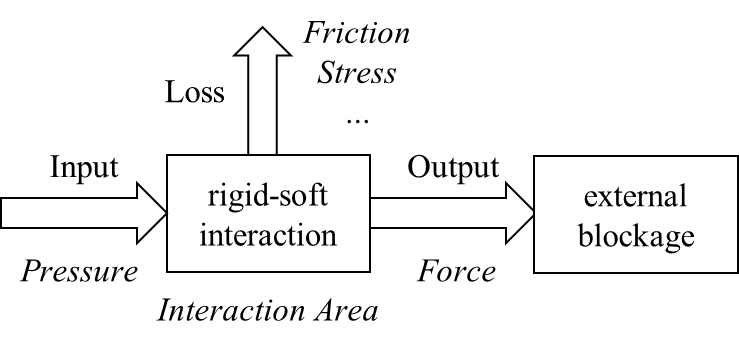}
\caption{Modeling the hybrid actuator during rigid-soft interaction, including the input pressure, output force, and energy loss during the nonlinear material deformation interaction with the rigid shells. Note that the soft actuator inside tends to fill all available volume inside the rigid shells under fluidic pressure.}
\label{fig:SystemDiagram}
\end{figure}

The exerted force from the interaction between the inflated balloon and the sliding shell can be directly measured by the loading cell, which is the system output as shown in Fig. \ref{fig:SystemDiagram}. One can compare the measured force as an ideal pressure-force-area model factored with a force loss parameter $\eta$. As shown in Eq. (\ref{eq:FPAeta}), $P_{in}$ is the input pressure, $A$ is the interaction area, and $F_{out}$ is the measured force exerted by the moving cylinder shell, which can be measured using a loading cell. The simplicity of the design enables one to model the actuator without dive into the non-linear mechanics of the soft actuator while effective characterizing the performance of the overall actuator in a light-weight design. 

\begin{equation}
\eta = 1 - \frac{F_{out}}{P_{in} \cdot A} \times 100\%
\label{eq:FPAeta}
\end{equation}

\begin{figure}[thpb]
\centering
\includegraphics[width=0.5\textwidth]{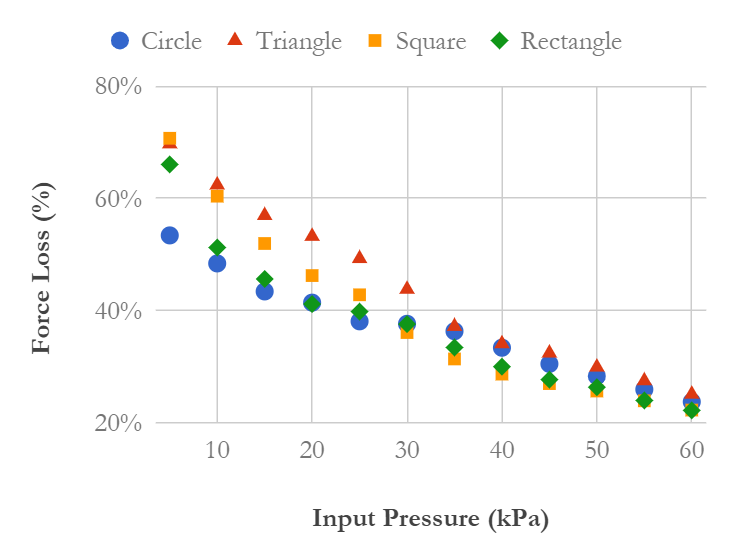}
\caption{The increasing efficiency of the hybrid actuators with different cross-section geometries while pressure increases.}
\label{fig:Efficiency}
\end{figure}

Four experiments are carried out to explore the characteristics of $\eta$ with different geometries of the interaction area using the same small, thin, and latex balloon, including circle, triangle, square and rectangle. All shapes of the interaction geometry shares the same area as a radius 25 mm circle. Such balloons can be cheaply sourced but their elasticity will be permanently changed once inflated. Therefore, the balloon is inflated and deflated continuously for 10 times before putting into the rigid shells for experiment. During each experiment, the pressure is supplied at an increment of 5 kPa until 60 kPa. We observed that given our design, further pressurization is at the risk of fraction the experiment platform with 3D printed plastic parts. At each pressure increment, the output force is measured three times. The averaged measurement is plotted in Fig. \ref{fig:Efficiency}, where the efficiency of the force exerted, $\eta$, increases as the pressure increases. 

As shown in Fig. \ref{fig:Efficiency}, the $\eta$ started with a relatively low value between 30\% and 50\% for different interaction geometries at the beginning. After 30 kPa, the $\eta$ starts to emerge with a linear relationship to the input pressure, reaching about 77\% at 60 kPa for all shapes. One possible explanation for the behavior before 30 kPa is the pre-pressurization of the balloon to fill all available cavity of the rigid shells with different interaction area. After this pre-pressurization, the overall actuator increase with a tendency towards a hydraulic cylinder with decreasing loss of force exerted. 

Among the four different interaction geometries tested, it is observed that the square and rectangle ones exhibited a relatively lower force loss at a similar level. The square one was found to be the relatively low performing one, with the circular one performs in the middle. This provides design guidelines for the soft actuator design when replacing the balloon with custom made soft actuators, where a square or rectangle with rounded corners might be a preferred choice for improved performance, which will be further addressed in the next section.

\begin{figure}[thpb]
\centering\includegraphics[width=0.5\textwidth]{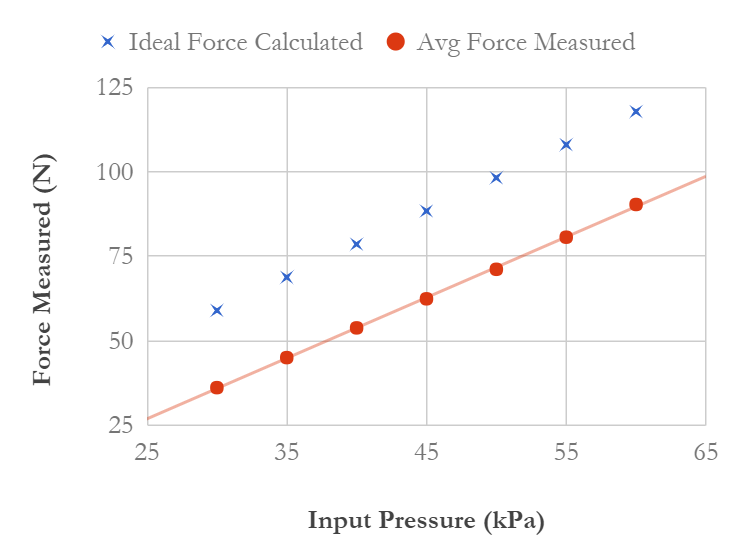}
\caption{The comparison between the ideal and average measured force exerted by the hybrid actuator.}
\label{fig:ForceCompare}
\end{figure}

The average force exerted by the actuator between 30 kPa and 60 kPa ranges between 36 N and 90 N, which is significantly larger than soft actuators made from purely soft material or the reinforced ones. The average force loss of the actuator between 30 and 60 kPa can be described using Eq. (\ref{eq:etaFit}) by fitting the measured data, reaching a high $r^2$ of 97.8\%. As shown in Fig. \ref{fig:ForceCompare}, the exerted force within a range of relatively lower pressure can reach up to around 90 N at 60 kPa, which is much higher than most actuators made from purely soft material. This enables the possibility of adopting such light-weight, low-cost and high performance hybrid actuator design to be implemented for lower limb wearable devices, which will be introduced in the next section. 

\begin{equation}
\eta = -0.005 \cdot x +0.522
\label{eq:etaFit}
\end{equation}

%%%%%%%%%%%%%%%%%%%%%%%%%%%%%%%%%%
\section{Lower-limb Wearable for Gait Augmentation}
%%%%%%%%%%%%%%%%%%%%%%%%%%%%%%%%%%

An engineering design of the proposed actuator is developed by using soft actuator instead of balloons for robust, reliable and improved performance, which can be integrated for lower-limb augmentation. Commonly gait abnormalities are caused by the misalignment of the femur, patella and the tibia, others may be caused by nerve system. These pathological gaits ultimately need surgical operations to rectify. However, some mild gait issues such as patients suffering from patellofemoral arthritis can alleviate the pain through applying medical wearable devices such as knee braces and supports. 

Commonly any misalignment of the lower limp can be rectified by exerting an external corrective moment through imposing 3-4 forces system along the side of the leg. To exert this force, many different styles of knee braces have been developed. Most braces offer similar aspects such as mechanical support, lateral stability, proprioception and protection for damaged ligaments. They also typically work to change the system of external forces and moments about a joint, as well as restore the balance between the external and internal moments \cite{Karavas13}. Common lower-limb assistance is usually powered by electric motors \cite{Walsh06}. Recent work in \cite{Robertson16} shows that it is possible to pack multiple soft actuators to generators high force output. 

\begin{figure}[thpb]
\centering
\includegraphics[width=0.5\textwidth]{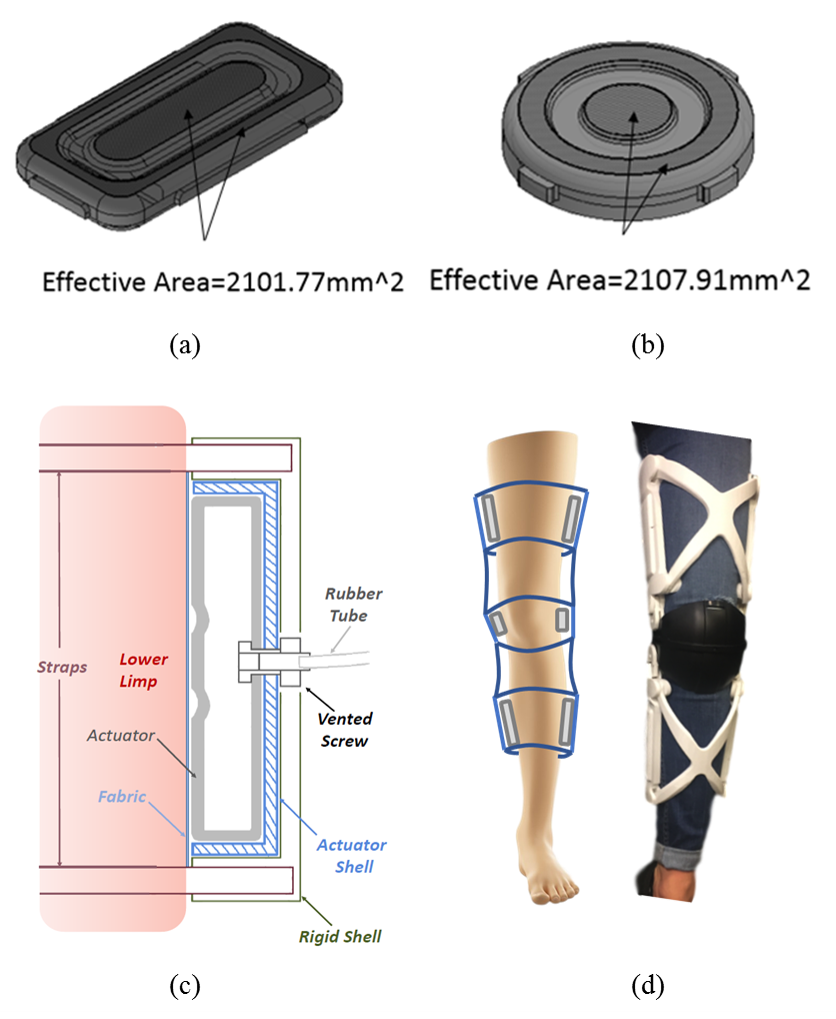}
\caption{Engineering design of the hybrid actuator for lower limb wearable as a gait augmentation device. (a) and (b) are two different interaction geometry with the same effective area. (c) is an illustration of the revised interaction mechanism of the hybrid actuator to produce compression against the leg. (d) is an illustration of overall wearable design with two 'X' structure to mount a total of six actuators on two sides of each leg. }
\label{fig:WearableActuator}
\end{figure}

There are a few design challenges for lower-limb wearable, including safe, compliant and powerful actuation, light-weight and flat design, ergonomics as a wearable device with active control. We adapt the principals of the proposed hybrid actuator design to address these design challenges by redesigning the actuator with a light-weight brace structure as shown in Fig. \ref{fig:WearableActuator}. The interaction area takes the geometry of a rectangle with rounded corners, which intends maximize the efficiency of the soft-rigid interaction under fluid pressurization. The soft actuator is replaced by a customized one fabricated through two stage molding with an increased thickness of 3 mm to safely hold the pressure inside. Furthermore, we removed the rigid, sliding shell by introducing a pop-up area on the top of the soft actuator so that the compression force is generated directly by the soft component that is contact with the leg. This enables us to reduce the overall thickness of the actuator to be the same as the mounting shell at the bottom, which is to be fixed to the bracelet structure. As a result, the modified actuator presents a compact form factor that works under a relatively lower input pressure and a small stroke distance to provide a compression force to the human leg for active augmentation. The overall bracelet structure takes an ``X'' shape that wraps around the leg, which can be individually customized during the design stage. Each bracelet comprises of two ``X'' structures tightened to the thigh and leg side of the knee. One actuator is placed on each side of the thigh, knee and leg. According to different augmentation need, different pairs of the actuators could be selectively pressurized to augment the alignment of the knee during a walking cycle or at rest. 

\begin{figure}[thpb]
\centering
\includegraphics[width=0.5\textwidth]{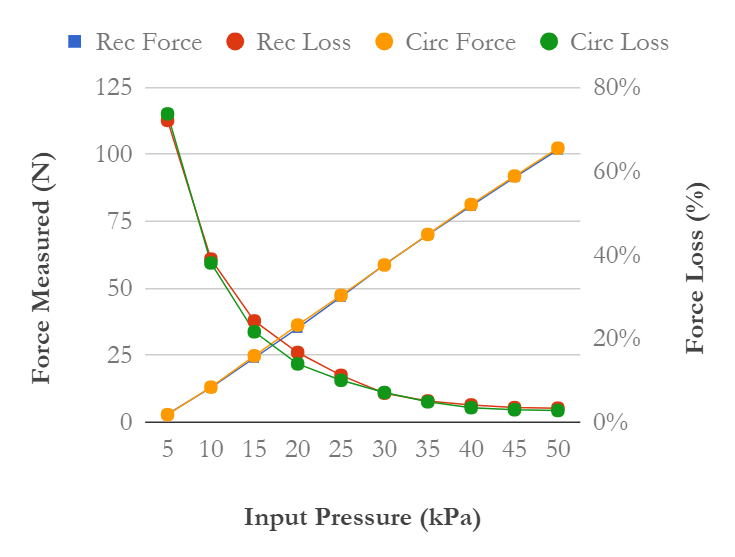}
\caption{Measured force output of the engineering designed hybrid actuator with different interaction geometry but similar output, where a high force of more than 100N can be generated at a low input pressure of 50 kPa with a very small loss of force (3\%) comparing to the ideal fluidic cylinder model.}
\label{fig:ModifiedPerformance}
\end{figure}

We also conducted experiments to characterize the performance of the modified hybrid actuator, as shown in Fig. \ref{fig:ModifiedPerformance}. The force output between the two different cross-sectional geometries are almost identical, where the exerted forces exhibit a linear relationship to the input pressure. The loss of force decrease exponentially from 70\% to about 3\% at 50 kPa. In fact, the force exerted from the actuator is more than enough for the wearable to be effective, which suggest a further design iteration to reduce the size and output within a more appropriate range as a lower-limb wearable. 

Each brace is equipped with six actuators, with three of them placed on each side of the leg. One pair is placed on the two sides of the knee, one pair on the thigh and one pair on the leg. To correct the knee alignment, different actuators are engaged at each phase of a gait cycle, aiming at correcting the alignment angle during active walking exercises. The interacting surface is through direct contact with the soft actuator, which is safe and compliant as a wearable device. Further development is currently on-going for a control system of this wearable device during active walking.

%%%%%%%%%%%%%%%%%%%%%%%%%%%%%%%%%%
\section{Conclusion}
%%%%%%%%%%%%%%%%%%%%%%%%%%%%%%%%%%

In this paper, we proposed the conceptual design of a fluidic soft actuator with rigid shells that is capable of generator large force at a lower operating pressure. We experimented with a quick, easy and cheap validation method by using latex party balloons as replacement of soft actuators for initial design exploration. We developed an engineering design of the actuator for a lower-limb wearable to actively correct the knee alignment angle in a light-weight structure. The resultant actuator performs very well with more than 100 N force exerted at only 50 kPa with only 3\% loss of force given the input pressure and actuator design. 

There are still many limitations with our proposed design. For example, the current engineering design generated too much force, giving us the design space to further reduce the size of the actuator as a wearable device. We are currently in the process of developing a control system for the wearable, and a user interface for more intuitive usage and operation. A portable version of the pneumatic power source is also to be developed to make this device wearable for daily usage.

\addtolength{\textheight}{-12cm}   % This command serves to balance the column lengths


%%%%%%%%%%%%%%%%%%%%%%%%%%%%%%%%%%
\section*{ACKNOWLEDGMENT}
%%%%%%%%%%%%%%%%%%%%%%%%%%%%%%%%%%
This work was supported in part by the Science, Technology, and Innovation Committee of Shenzhen City [JCYJ20160422145322758].

\end{document}